\pdfoutput=1
\documentclass[11pt]{article}
\usepackage{acl}

\usepackage{times}
\usepackage{latexsym}
\usepackage[T1]{fontenc}
\usepackage[utf8]{inputenc}

\usepackage{microtype}

\usepackage{graphicx}

\usepackage{amsmath}
\usepackage{subcaption}

\title{M2K-VDG: Model-Adaptive Multimodal Knowledge Anchor Enhanced Video-grounded Dialogue Generation}

\author{Hongcheng Liu $^1$, Pingjie Wang$^1,2$, Yu Wang$^{*1,2}$, Yanfeng Wang$^{1,2}$ \\
  $^{1}$Cooperative Medianet Innovation Center, Shanghai Jiao Tong University \\
  $^{2}$Shanghai Artificial Intelligence Laboratory \\
  \texttt{\{hongcheng\_liu,pingjiewang,yuwangsjtu,wangyanfeng622\}@sjtu.edu.cn} \\
}

\begin{document}
\maketitle

\begin{abstract}
Video-grounded dialogue generation (VDG) requires the system to generate a fluent and accurate answer based on multimodal knowledge. However, the difficulty in multimodal knowledge utilization brings serious hallucinations to VDG models in practice. Although previous works mitigate the hallucination in a variety of ways, they hardly take notice of the importance of the multimodal knowledge anchor answer tokens. In this paper, we reveal via perplexity that different VDG models experience varying hallucinations and exhibit diverse anchor tokens. Based on this observation, we propose M2K-VDG, a model-adaptive multimodal knowledge anchor enhancement framework for hallucination reduction. Furthermore, we introduce the counterfactual effect for more accurate anchor token detection. The experimental results on three popular benchmarks exhibit the superiority of our approach over state-of-the-art methods, demonstrating its effectiveness in reducing hallucinations.

\end{abstract}
\section{Introduction}

\begin{figure}[t]
\setlength{\belowcaptionskip}{-0.5cm}
    \centering
    \includegraphics[width=0.85\linewidth]{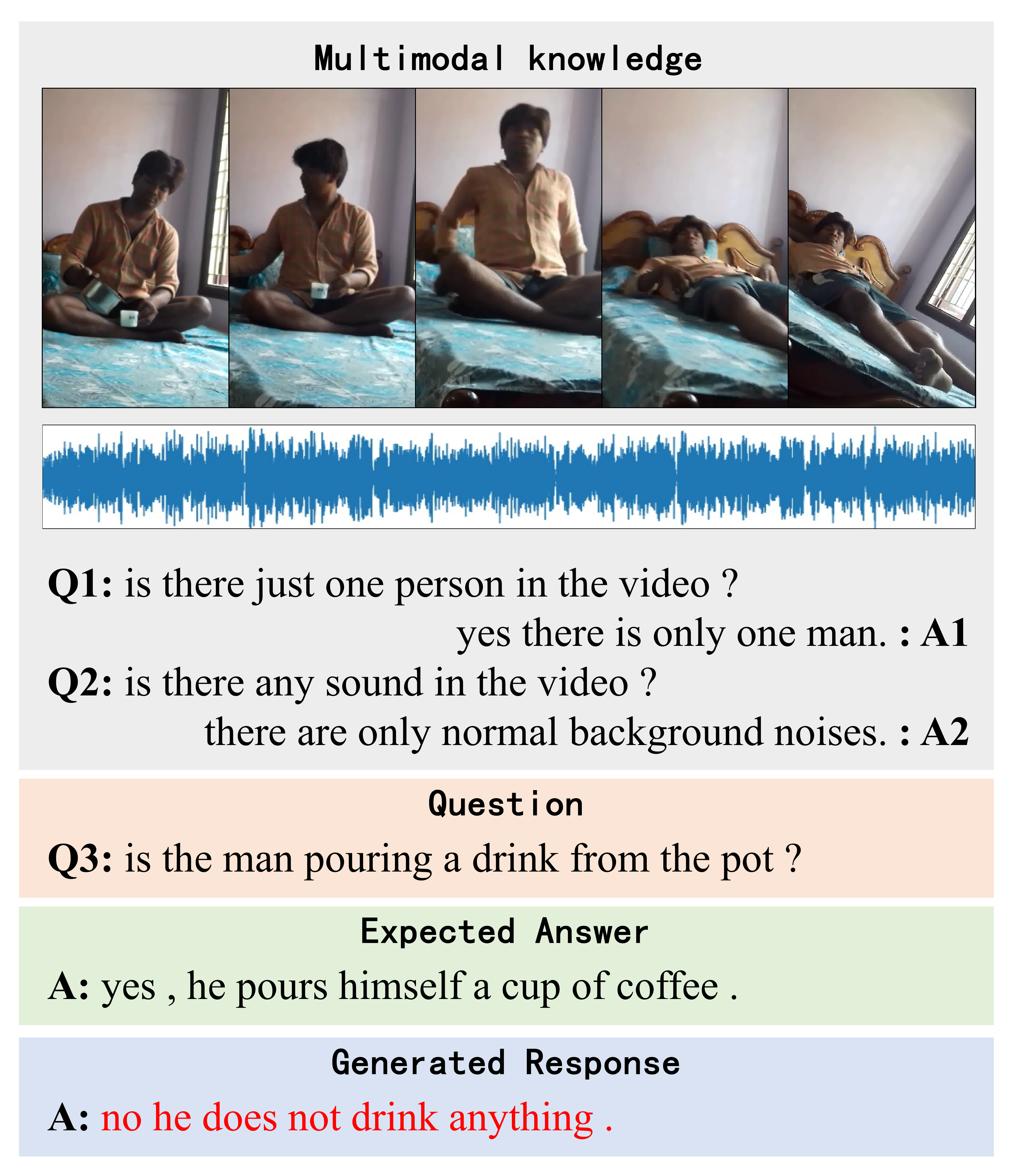}
    \caption{The demonstration of the video-grounded dialogue generation task, where the system is required to generate a fluent and accurate response according to the multimodal knowledge. However, the difficulty in multimodal knowledge utilization always leads to the experience of the hallucination.}
    \label{fig:VDG}
\end{figure}

The task of video-grounded dialogue generation (VDG) involves generating responses to queries by leveraging multimodal knowledge sources \cite{avsd}, including visual content, auditory features, and previous dialogue exchanges, as 
illustrated in Figure~\ref{fig:VDG}. This task stands out from traditional knowledge-based question-answering challenges due to the intricate nature of its knowledge sources. The integration of diverse modalities introduces a level of complexity not typically encountered in other domains, making effective utilization of this information a formidable challenge \cite{MSG-BART}. Consequently, VDG systems are more prone to generating inaccurate or irrelevant responses, which is a phenomenon known as `hallucination' in the realm of large language models.
To reduce hallucination and enhance performance, there are various methods proposed to address this issue, such as external video description and various dataset enhancement \cite{av-trans, le-etal-2023-c3}. Besides, \citet{cherian20222} and \citet{geng2021dynamic} utilize symbolic graphs for multimodal alignment. However, these approaches only focus on more inputs but ignore reducing the hallucination specifically based on its characteristics. 

\begin{figure}[t]
\setlength{\belowcaptionskip}{-0.6cm}
    \centering
    \includegraphics[width=0.9\linewidth]{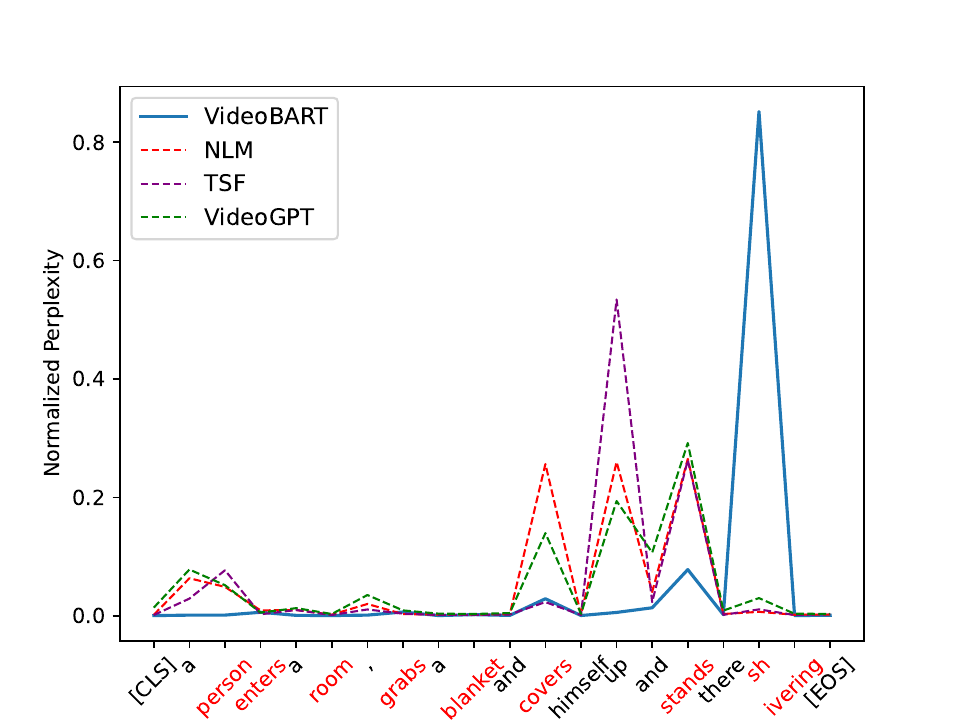}
    \caption{The perplexity of answer tokens derived by different VDG models trained with the same multimodal content, and the higher perplexity means the more serious hallucination. The question is `\textbf{tell me the sequence in which the event occurred?}`, and the label in red denotes the knowledge-related tokens of ground answers through human detection. It is noted that different models experience hallucinations in various anchor tokens, which reflects the various multimodal knowledge anchors among different models.}
    \label{fig:ppl}
\end{figure}

To analyze the hallucination in the VDG, we evaluated it across different models using perplexity. This measurement compares the generated probability and the ground answer, and is commonly used to assess the quality of generated results \cite{ppl_commen_metric}. As shown in Figure~\ref{fig:ppl}, the hallucination varies among tokens within a sentence, with certain tokens being more susceptible to hallucination. In this paper, we refer to these tokens as `anchor tokens', representing key tokens in reducing hallucination. Furthermore, these anchor tokens are closely linked to multimodal knowledge, hence, we categorize them as `multimodal knowledge anchor tokens'. It can be seen in Figure~\ref{fig:ppl} that the degree of perplexity notably varies across models, and different models tend to hallucinate more on different multimodal knowledge anchor tokens. Given the significant degree of hallucination caused by anchor tokens, it is important to use these tokens effectively to reduce hallucinations.

Considering the importance of the anchor tokens, FocusL~\cite{deng-etal-2023-towards} forces the model to be more attentive to knowledge anchor tokens and makes great progress in hallucination mitigation. Nevertheless, this method employs cosine similarity between the answer tokens and given text knowledge to select anchor tokens, regarding the anchor tokens as the immutable standards among various models and overlooking the varying knowledge anchor tokens of different methods as illustrated in Figure~\ref{fig:ppl}. 
Moreover, due to semantic gaps, this method cannot be applied to video-grounded dialogue generation directly.


To address the aforementioned issues, we propose M2K-VDG, a \textbf{M}odel-adaptive \textbf{M}ultimodal \textbf{K}nowledge anchor enhancement framework for the \textbf{VDG} task. This framework reduces model hallucinations by enhancing multimodal knowledge anchor tokens from the model itself. 
Based on the observation in Figure~\ref{fig:ppl}, we split our M2K-VDG approach into two stages: 1) Anchor Token Detection, and 2) Model Hallucination Reduction. In the first stage,
we utilize the perplexity as the anchor weight of each answer token derived by well-trained models for model adaptive anchor detection.
In the second stage, the derived anchor weight is then normalized and merged into the original Softmax function to enhance 
training for knowledge anchors. To further enhance the precision and resilience of our anchor detection process, we introduce an innovative method based on counterfactual effects. This method provides a more nuanced assessment of anchor significance by examining the hypothetical impact of altering the inputs. Through this approach, we aim to refine the detection and utilization of knowledge anchors within the M2K-VDG framework, ensuring a more reliable and effective reduction in model hallucinations.

To evaluate the performance of M2K-VDG, we compare it with a range of state-of-the-art methods on three popular VDG benchmarks: AVSD10 dataset,  NExT-OE dataset, and MUSIC-AVQA dataset. Experimental results verify that M2K-VDG exhibits significantly superior performance compared to established baseline methodologies across all benchmarks, demonstrating the superiority and effectiveness of our proposed 
multimodal knowledge anchor enhancement approach.

Our main contributions are three folds: 

\begin{enumerate}
\item [$\bullet$]\textbf{Diverse Hallucination Patterns Across VDG Models}: Our research uncovers that various VDG models exhibit different patterns of hallucinations. To address this issue, we introduce M2K-VDG, a novel framework specifically designed to adaptively reduce hallucinations. This is achieved through two core strategies: Detection of Anchor Tokens and Reduction of Model Hallucinations, which together aim to improve model accuracy and reliability by anchoring generated content to detectable knowledge points within the model's output.
\item [$\bullet$]\textbf{Enhanced Anchor Detection with Counterfactual Analysis}: We advance the robustness of our framework's anchor detection mechanism by incorporating a counterfactual effect. This method significantly improves our ability to detect multimodal knowledge anchor tokens, offering a more effective means of identifying critical information points that ground the model's output in reality.
\item [$\bullet$]\textbf{Comprehensive Evaluation on Popular VDG Benchmarks}: We rigorously test M2K-VDG against three widely recognized VDG benchmarks to validate its effectiveness and efficiency. The results clearly demonstrate that M2K-VDG surpasses existing state-of-the-art methods, showcasing its superiority in reducing model hallucinations and enhancing the overall quality of visual dialog generation. This underscores the potential of M2K-VDG to set a new benchmark in the field, offering a robust solution to the longstanding challenge of model hallucination in VDG tasks.
\end{enumerate}

\section{Related Works}
Video-grounded dialogue generation (VDG) requires the system to generate answers related to the questions based on multimodal information~\cite{AVSD-SIMPLE}, which consists of video, audio, and dialogue history. Compared with other visual-language tasks, the VDG exhibits significant difficulty in the utilization of multimodal knowledge consisting of the visual, auditory, and textual modalities~\cite{pasunuru-bansal-2018-game,MSG-BART}, which leads to serious hallucinations.

To more effectively reduce hallucination, a variety of methods have been proposed that better utilize different modalities. Several models minimize hallucination by enhancing the connection between visual features and questions. This is achieved through various feature extractors \cite{videogpt,timesformer-gpt}, external visual information \cite{2020-bist,jin2023knowledge}, and fusion strategies \cite{zhao-etal-2022-collaborative, MAG}. In addition, manifold methods are proposed for enhancing the utilization of auditory information~\cite{lao2023coca,yoon2023hear, Li_2022_CVPR}. Furthermore, there are many improvement strategies proposed for the textual modality, such as DialogMCF~\cite{DialogMCF} and SCGA~\cite{kim2021structured}. 
However, all the above approaches are short of analyzing the inherent characteristics of the hallucination, thereby attaining limited performance improvement. In this paper, we propose M2K-VDG to tackle this problem. Specifically, M2K-VDG detects the anchor tokens with more serious hallucinations and utilizes them to achieve higher answering quality.

\section{Preliminaries}
In this section, we introduce the basic concept of causal inference and counterfactual effect. For consistency, we use uppercase letters (e.g. $X$) to represent the variable and lowercase letters (e.g. $x$) for observed values.

\begin{figure}[t]
	\centering
	\begin{subfigure}{0.47\linewidth}
            \setlength{\belowcaptionskip}{-0cm}
		\centering
		\includegraphics[width=0.9\linewidth]{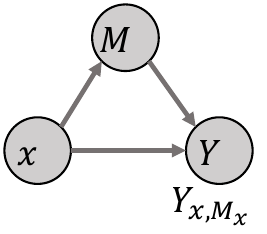}
		\caption{$Y_{x,M_x}$}
		\label{fig:causal_graph1}
	\end{subfigure}
	\centering
	\begin{subfigure}{0.47\linewidth}
            \setlength{\belowcaptionskip}{-0cm}
		\centering
		\includegraphics[width=0.9\linewidth]{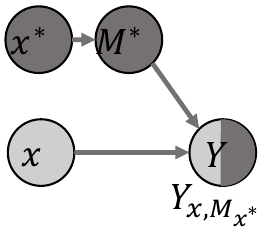}
		\caption{$Y_{x,M_{x^*}}$}
		\label{fig:causal_graph2}
	\end{subfigure}

        \begin{subfigure}{0.47\linewidth}
            \setlength{\belowcaptionskip}{-0cm}
		\centering
		\includegraphics[width=0.9\linewidth]{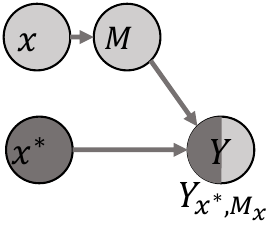}
		\caption{$Y_{x,M_{x^*}}$}
		\label{fig:causal_graph3}
	\end{subfigure}
	\centering
	\begin{subfigure}{0.47\linewidth}
            \setlength{\belowcaptionskip}{-0cm}
		\centering
		\includegraphics[width=0.9\linewidth]{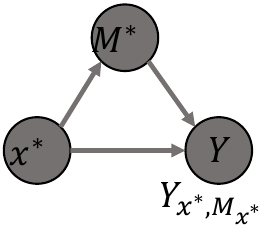}
		\caption{$Y_{x^*,M_{x^*}}$}
		\label{fig:causal_graph4}
	\end{subfigure}
	\caption{The illustration of the causal effect.}
	\label{fig:causal_graph}
\end{figure}

\subsection{Causal Graph} 

The causal connections among distinct variables are encapsulated in a causal graph $\mathcal{G} = \{\nu,\varepsilon \}$, where $\nu$ signifies the set of variables, and $\varepsilon$ characterizes the causal relationships existing between them. 
The example in Figure~\ref{fig:causal_graph} illustrates the causality between variable $X$, mediator $M$, and variable $Y$.The $X \to Y$ denotes variable $X$ has a direct causal effect on variable $Y$, and the $X \to M \to Y$ denotes variable $X$ has an indirect causal effect on variable $Y$ via mediator $M$. 
\subsection{Counterfactual notations}
For computational reasoning, the causal graph can be formulated as causal notations, which is
\begin{equation}
\small
    Y_{x,m} = Y(do(X=x),M=m) = Y(X=x,M=m),
\end{equation}
where $m = M_x = M(X=x)$, the $do$ operator can be omitted without any confounder in $X$. Furthermore, the variable $X$ can be set as $x^*$ to represent the negation operation in the counterfactual reasoning, which is
\begin{equation}
\small
    Y_{x^*,m} = Y(X=x^*,M=m),
\end{equation}
where the negation operation can be achieved by omitting input $X$ in practice.

\begin{figure*}[t]
\setlength{\belowcaptionskip}{-0.5cm}
    \centering
    \includegraphics[width=0.9\linewidth]{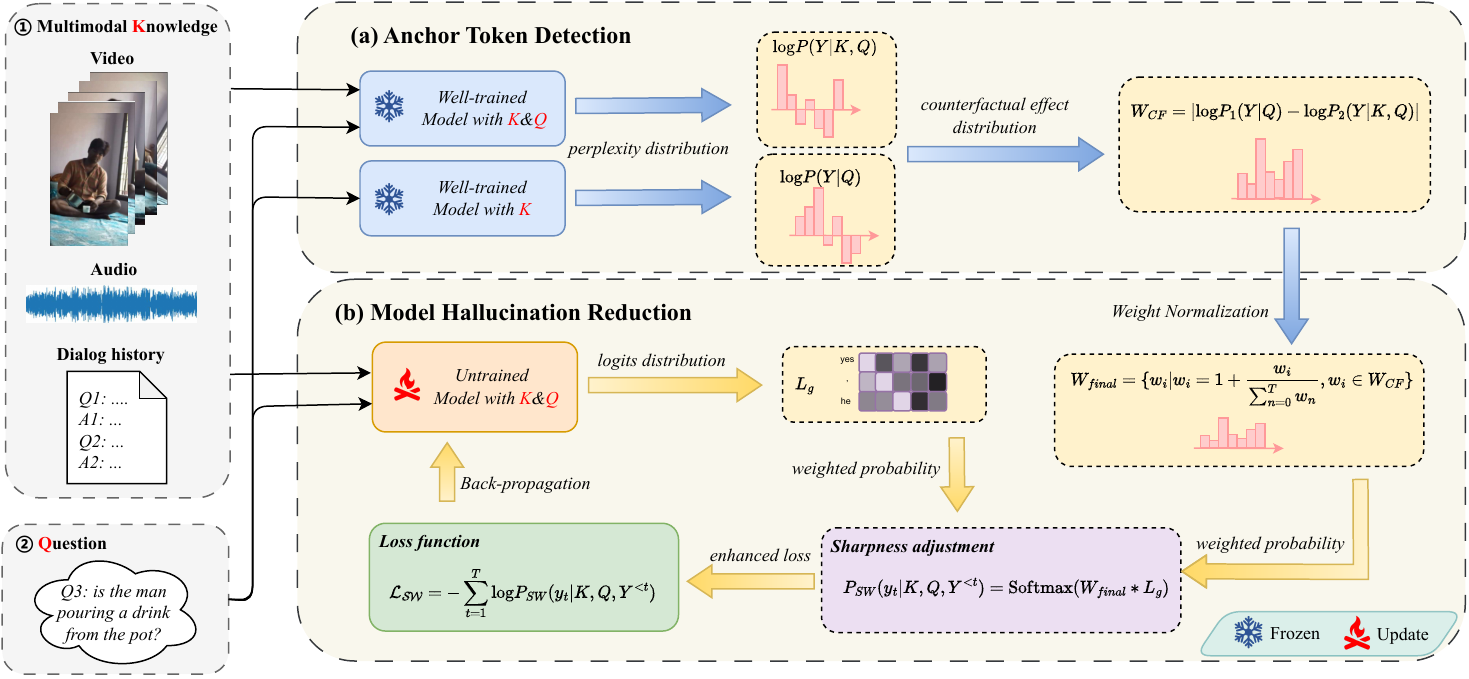}
    \caption{\textbf{The overview of the M2K-VDG}: (a) Anchor Token Detection aims to detect the multimodal knowledge anchor tokens in the grounded answer through well-trained models, and is categorized into perplexity- and counterfactual effect-based. (b) Model Hallucination Reduction normalizes the anchor weight and focuses on the new model training via temperature change derived by the normalized weight.}
    \label{fig:overview}
\end{figure*}

\subsection{Counterfactual Effect}
\label{sec:counter}
Counterfactual effects are often used to measure the impact of the input variable on the output variable, which can minimize other distractions. In practice, it often uses total effect (TE) to measure the impact through the difference in $Y$ in the condition of the different treatments on $X$, which is denoted as:
\begin{equation}
\small
    TE = Y_{x,M_x} - Y_{x^{*},M_{x^*}},
\end{equation}
where the TE can reflect the total impact of $X$ on $Y$. Additionally, the total effect can be further divided into $TIE$ and $NDE$, which can measure the impact on variables in a more fine-grained way. The formulations are

\begin{equation}
\small
    TIE = Y_{x,M_x} - Y_{x,M_{x^*}},
\end{equation}
\begin{equation}
\small
    NDE = Y_{x,M_{x^*}} - Y_{x^*,M_{x^*}},
\end{equation}
where the TIE can reflect the impact of $M$ on $Y$ and the NDE can reflect the effect of $X$ on $Y$ on condition of $M_{x^*}$. Considering it can reflect the impact of $X$ on $Y$ is highly related to the correlation between the $X$ and $Y$ without any other factors interference, the counterfactual effect is widely used for impact measurement and relevance calculation. 
Therefore, we utilize the counterfactual effect as the metric to measure the anchor degree of answer tokens with multimodal knowledge for the VDG task.
\section{Methods}
\subsection{Task Formulation}
Given the video $V$, audio $A$, dialogue history $H$, and question $Q$, the video-grounded dialogue generation aims to generate appreciative and fluent responses $Y=\{y_1,y_2,\dots,y_T\}$, which contains $T$ tokens. For conciseness, we use multimodal knowledge $K$ to represent the set of the video, audio, and dialogue history, that is, $K=\{V,A,H\}$.
This task can be formulated as
\begin{equation}
\small
    P(Y | K, Q) = \prod_{t=1}^{T} P(y_t| K, Q, Y_{<t}).
\end{equation}

\subsection{Overview}

The overview of the M2K-VDG is shown in Figure~\ref{fig:overview}, which is divided into two stages: 1) anchor token detection, and 2) model hallucination reduction. Specifically, for the first stage, we measure the perplexity of the answer tokens by well-trained models and derive the anchor degree of each token. The second stage aims to utilize the derived anchor degrees of tokens as a weight to enhance the untrained model learning process, in which the anchor degree is normalized and merged into the answer probability distribution and loss function.

We adopt the BART model as our backbone and extend it to adapt to multi-modality input, named VideoBART.
Specifically, we use the BART encoder as the multimodal knowledge extractor and the BART decoder as the response generator, as the encoder and decoder are suitable to be responsible for complex understanding and detailed reasoning. 
In detail, we feed the video, audio, and dialogue history into the BART encoder to obtain a multimodal representation, and then feed it into the BART decoder along with the question to generate the response.  Specifically, we use special tokens to separate different modalities.

\subsection{Anchor Token Detection}
\label{sec:weight}


As we have observed in Figure~\ref{fig:ppl}, VDG models are prone to experience hallucination for specific anchor tokens, which are correlated to multimodal knowledge and more easily to be falsely predicted. Furthermore, the anchor tokens have limited opportunity to be optimized as they are highly correlated to specific samples. To address the above issues, we first calculate the anchor degree of each answer token and then emphasize the tokens with higher anchor weights during training.

However, considering the huge gap in representations between natural language and other modalities, the traditional strategies for relevance calculation (e.g., cosine similarity) are unsuitable for the VDG task. 
To tackle this problem, we first leverage perplexity to detect the anchor degree of tokens for its high correlation with the hallucination probability. Furthermore, we improve the detection method by the counterfactual effect, which is more effective and robust.
It is noted that both strategies can adapt to and enhance various backbones, which we will demonstrate in Section~\ref{sec:abl}.

\subsubsection{Perplexity-based Detection}

Since the perplexity can efficiently identify anchor tokens from the model, as demonstrated in Figure~\ref{fig:ppl}, we use it to detect these tokens and determine the anchor weight for training.
The anchor weight $W_P$ derived by this perplexity-based detection method is formulated as
\begin{equation}
\small
    W_{P} =  -\log P\left(y_t \mid K, Q, Y_{<t}\right).
\end{equation}
It is noted that the anchor weight is obtained by a well-trained model, which can be any existing VDG model.
While the perplexity-based detection method can be effective, it is easily influenced by variations in noise during the training phase, thereby resulting in different anchors.  
Moreover, this method is unable to specifically detect multimodal knowledge anchor tokens. It instead represents the cumulative effect on different modality inputs, which imposes limitations in practice.


\subsubsection{Counterfactual Effect-based Detection}

To tackle these problems, we further introduce the counterfactual effect to enhance the robustness and effectiveness of anchor degree estimation. Specifically, as described in Section \ref{sec:counter}, we utilize two well-trained models for response generation under different conditions, one of which is trained given both multimodal knowledge and the question, and the other is only fed with the question. On this occasion, the counterfactual effect is denoted as
\begin{equation}
\small
    TIE = Y(K=k,Q=q) - Y(K=k^*,Q=q),
\end{equation}
We reformulate this effect in a perplexity form for the VDG task as
\begin{equation}
\small
    W_{CF} =  \left | \log P_1\left(y_t \mid Q, Y_{<t}\right)-\log P_2\left(y_t \mid K, Q, Y_{<t}\right) \right |,
    \label{equ:w_cf}
\end{equation}
where $\left | \cdot \right |$ denotes the absolute value, $P_1$ and $P_2$ represent the perplexity distributions of the two models trained with different inputs. In this way, we avoid the estimation noise in the training stage. Additionally, this method can also be extended to measure the anchor degree of tokens for a specific modality by controlling the input modality to $P_1$.

\begin{table*}[h]
\setlength{\belowcaptionskip}{-0.5cm}
\centering
\fontsize{10.5pt}{\baselineskip}\selectfont
\begin{tabular}{lccccccc}
\hline
\textbf{Methods} & \textbf{BLEU1}& \textbf{BLEU2}& \textbf{BLEU3}& \textbf{BLEU4}& \textbf{METEOR}& \textbf{ROUGH-L}& \textbf{CIDEr}\\ \hline
AV-trans~\citeyearpar{av-trans} & - & - & - & 0.247     & 0.191     & 0.437     & 0.566    \\
Ext.AV*~\citeyearpar{av-trans} & - & - & - &0.371 &0.245 &0.535 &0.869 \\
NLM~\citeyearpar{med-cat}    & {0.641}  & {0.489}  & {0.379}  & {0.298}  & {0.225}  & {0.502}  & {0.804}  \\
MED-CAT~\citeyearpar{med-cat}& 0.670& 0.541& 0.441&0.365& 0.241 &0.526 &0.906     \\
UniVL-obj*~\citeyearpar{med-cat}&0.673 &0.545 &0.448& 0.372       & 0.243       & 0.530        & 0.912      \\
TSF(ensemble)~\citeyearpar{timesformer-gpt}   & 0.680     & 0.558     & 0.461     & 0.385     & 0.247     & 0.539     & 0.957     \\
DialogMCF~\citeyearpar{DialogMCF}& 0.693     & 0.556     & 0.450     & 0.369     & 0.249     & 0.536     & 0.912     \\
MSG-BART*~\citeyearpar{MSG-BART} & -     & -    & -     & 0.390    & 0.268    & 0.556     & 1.008     \\
\textbf{M2K-VDG (Ours)} & {\textbf{0.723}} & {\textbf{0.589}} & {\textbf{0.483}} & {\textbf{0.398}} & {\textbf{0.271}} & {\textbf{0.560}} & {\textbf{1.020}} \\ \hline
\end{tabular}
\caption{Evaluation results of our model compared with baseline approaches on AVSD10 dataset. The * denotes the result obtained with the external knowledge, including the detected objects, scene graphs, and video descriptions.
}
\label{tab:main_avsd}
\end{table*}


\subsection{Model Hallucination Reduction}
As we have obtained the anchor weight for each answer token, we apply normalization and merge it into the original all-1 vector to obtain the final anchor weight of tokens. Then, we utilize the final weight in the Softmax function to reduce the multimodal knowledge hallucination by forcing the model to be more attentive to anchor tokens.

\subsubsection{Weight Normalization}

Although these two detection strategies can both measure the anchor degree of the answer token effectively, they cannot be applied for hallucination reduction directly due to their unequal ranges. Therefore, we normalize the obtained weight and merge it into the original all-1 vector, which is formulated as
\begin{equation}
\small
    W_{final} = \{ w_t |w_t = 1 + \frac{w_t}{\sum_{n=0}^{T}w_n }, w_t \in W \},
\end{equation}
where $W$ denotes the $W_P$ or $W_{CF}$, and $T$ represents the total number of tokens within a sentence.


\subsubsection{Anchor Enhanced Loss}
After obtaining the final weight, we adopt it as the temperature rate in the Softmax function to adjust the sharpness of each answer token, which is formulated as
\begin{equation}
\small
    P_{SW}(y_t| K, Q, Y_{<t}) = \rm{Softmax}(W_{final}*L_g),
\end{equation}
where $L_g$ denotes the output logits distribution. The modified probability $P_{SW}$ is then utilized to enhance the loss function as
\begin{equation}
\small
    \mathcal{L_{SW}}=-\sum_{t=1}^{T} \log P_{SW}\left(y_t \mid K, Q, Y_{<t}\right).
\end{equation}

\section{Experiments}
\subsection{Datasets}

Experiments are conducted on three popular VDG benchmarks, which contain AVSD10, NExT-OE, and MUSIC-AVQA datasets. These datasets are different in scenes, modalities, and fields. Furthermore, we adopt different evaluation metrics from various aspects, including fluent expression (BLEU1-4, METEOR, ROUGH-L, CIDEr), semantic representation (WUPS Score\footnote{Wu-Palmer Similarity Score: evaluate the semantic similarity between the generated answers and the ground truth answers, which are available in the dataset.}), and accurate generation (Accuracy). The details of the datasets are given in Appendix~\ref{sec:appendix1}.

\begin{table}[t]
\setlength{\belowcaptionskip}{-0.5cm}
\centering
\begin{tabular}{lc}
\hline
\multicolumn{1}{c}{\textbf{Methods}}& \textbf{WUPS} \\
\hline
HCRN~\citeyearpar{le2020hierarchical}    & 23.92 \\
HME~\citeyearpar{fan2019heterogeneous}     & 24.06 \\
UATT~\citeyearpar{xue2017unifying}     & 24.25 \\
HGA~\citeyearpar{jiang2020reasoning}     & 25.18 \\
ClipBERT~\citeyearpar{lei2021less}& 24.17 \\
MSG-BART*~\citeyearpar{MSG-BART} &27.45 \\
KcGA*~\citeyearpar{jin2023knowledge}    & 28.20  \\
\textbf{M2K-VDG(Ours)}& \textbf{29.16} \\
\hline
\end{tabular}
\caption{The WUPS scores of our model compared with baseline approaches on the NExT-OE dataset. The * denotes the result obtained with external knowledge, including the knowledge base and sence graph.}
\label{tab:main_next}
\end{table}

\begin{table*}[t]
\setlength{\belowcaptionskip}{-0.5cm}
\centering
\fontsize{10.0pt}{\baselineskip}\selectfont
\begin{tabular}{lcccccccccc}
\hline
\textbf{Methods} & \rm{\textbf{ $A_{C}$}}  & \rm{\textbf{  $A_{Cm}$}}  & \rm{\textbf{$V_{C}$}}  & \rm{\textbf{$V_{L}$}}  & \rm{\textbf{$AV_{E}$}}  & \rm{\textbf{$AV_{C}$}}  & \rm{\textbf{$AV_{L}$}}  & \rm{\textbf{$AV_{Cm}$}} & \rm{\textbf{$AV_{T}$}} & \textbf{Average}\\ \hline
CONV$\dag$~\citeyearpar{9145807} & 74.07 &68.89 & 67.47 &54.56 & 82.91 &50.81 &63.03 &60.27 &51.58 & 63.65 \\
MCAN~\citeyearpar{8953581} &77.50 &55.24 & 71.56 &70.93 & 80.40 &54.48 &64.91 &57.22 &47.57 & 65.49 \\
HCRN~\citeyearpar{le2020hierarchical}   &68.59 &50.92 & 64.39 &61.81 & 54.47 &41.53 &53.38 &52.11 &47.69 & 55.73 \\
HME~\citeyearpar{fan2019heterogeneous}   &74.76 &63.56 & 67.97 &69.46 & 80.30 &53.18 &63.19 &62.69 &59.83 & 66.45 \\
AVSD~\citeyearpar{AVSD-SIMPLE}   &72.41 &61.90 & 67.39 &74.19 & 81.61 &58.79 &63.89 &61.52 &61.41 & 67.44 \\
P-AVQA$\dag$~\citeyearpar{Pano-AVQA}  &74.36& 64.56 & 69.39 &75.65 & 81.21 &59.33 &64.91 &64.22 &63.23 & 68.93 \\
AVST~\citeyearpar{Li_2022_CVPR}   &78.18 &67.05& 71.56 &76.38 & 81.81 &64.51 &\textbf{70.80} &\textbf{66.01} &63.23 & 71.52 \\
COCA~\citeyearpar{lao2023coca}    &79.94 &\textbf{67.68} & 75.10 &75.43 & \textbf{83.50} &66.63 &69.72 &64.12 &65.57   & 72.33\\
\textbf{M2K-VDG(Ours)} &\textbf{82.39}	&65.15	&\textbf{78.11}	&\textbf{78.12}	&82.49	&\textbf{70.75}	&64.13	&62.49	&\textbf{66.54} & \textbf{72.87}\\ \hline
\end{tabular}
\caption{The accuracy of our model compared with baseline approaches on the MUSIC-AVQA dataset. (A: audio, V: visual, AV: audio-visual, C: Counting, Cm: Comparative, L: Location, E: Existential, T: Temporal). The $\dag$ denotes the abbreviation of the method for convenient display. (CONV: CONVLSTM, P-AVQA: Pano-AVQA)}
\label{tab:main_music}
\end{table*}

\begin{table}[t]
\setlength{\belowcaptionskip}{-0.6cm}
\centering
\begin{tabular}{lcc}
\hline
\multicolumn{1}{c}{\textbf{Methods}}   & \textbf{BLEU4}     &\textbf{CIDEr}\\ \hline
VideoBART  & 0.384   & 0.987 \\
$W_P$   & 0.388 & 1.002\\
$W_{CF}$ & \textbf{0.398 } & \textbf{1.020}  \\ \hline
\end{tabular}
\caption{Evaluation results of detection methods ablation experiments on AVSD10 dataset. }
\label{tab:abl-formulations}
\end{table}

\begin{table*}[t]
\setlength{\belowcaptionskip}{-0.5cm}
\centering
\begin{tabular}{lccccccc}
\hline
\textbf{Methods}   & \textbf{BLEU1} & \textbf{BLEU2} & \textbf{BLEU3} & \textbf{BLEU4} & \textbf{METEOR} & \textbf{ROUGH\_L} & \textbf{CIDEr} \\ \hline
AV-trans  & -     & -     & -     & 0.247 & 0.191  & 0.437    & 0.566 \\
AV-trans+M2K        &0.606	&0.469	&0.354	&0.276	&0.201	&0.466	&0.629 \\ \hline
NLM      & 0.641	&0.489	&0.379 	&0.298 	&0.225 	&0.502	&0.804 \\
NLM+M2K    & 0.699	&0.563	&0.458	&0.376	&0.255	&0.543	&0.964 \\ \hline
TSF(ensemble)    & 0.680	&0.558	&0.461	&0.385&	0.247	&0.539	&0.957 \\
TSF+M2K & 0.675	&0.555	&0.462	&0.389	&0.246	&0.537	&0.957 \\ \hline
VideoGPT* & 0.682 & 0.559 & 0.458 & 0.380 & 0.245  & 0.537    & 0.937 \\
VideoGPT+M2K        & 0.700	&0.566	&0.461	&0.380	&0.255	&0.543	&0.970 \\ \hline
VideoBART      & 0.707 & 0.571 & 0.466 & 0.384 & 0.262  & 0.545    & 0.987 \\
VideoBART+M2K    & 0.723 & 0.589 & 0.483 & 0.398 & 0.271 & 0.560 & 1.020 \\ 
\hline
\end{tabular}
\caption{Evaluation results of baseline ablation experiments on AVSD10 dataset.
The * denotes the results we reproduced for the result missing and the M2K is an abbreviation of the multimodal knowledge anchor enhancement.
}
\label{tab:abl-methods}
\end{table*}

\subsection{Experimental Settings}
We extract visual features by ActionCLIP ~\cite{wang2021actionclip} and obtain auditory features by Wav2CLIP~\cite{wav2clip} with a time duration of 4s.
In the training phase, we choose $W_{CF}$ as the final anchor token detection method. We initialize our model by BART-base \footnote{\url{https://huggingface.co/facebook/bart-base}} and adopt an AdamW optimizer with a learning rate of 8.75e-5. The batch size is 32 and the smooth rate in Equation~\ref{equ:w_cf} is 0.1. During the inference phase, we used the beam search algorithm with a beam size of 6 and a penalty factor of 0.6.
\subsection{Baselines}
\label{Sec:baselines}
We use the following methods as our baseline system. (i) \textbf{VideoGPT}~\cite{videogpt} integrates visual-audio features and the dialogue text as a combined multimodal 
input
and feeds it into the GPT-2 to generate a response. Furthermore, this is the champion in the AVSD8 challenge and we reproduce it by official code for comparison. (ii) \textbf{TSF}~\cite{timesformer-gpt} employs TimeSformer as the visual feature extractor and enhances performance through ensemble methods. 
(iii) \textbf{AV-trans}~\cite{av-trans} uses the encoder-decoder structure as the backbone without pre-training and introduces bimodal attention to fuse audio-visual features.
(iv) \textbf{NLM}~\cite{med-cat} identifies actions as video features, transmitting them to GPT-2 for generating responses based on text.

\subsection{Main Result}

The main experimental results are shown in Table~\ref{tab:main_avsd}-\ref{tab:main_music}. It can be seen that our M2K-VDG methods outperform other baselines across almost all metrics, indicating their effectiveness in reducing hallucination. As shown in Table~\ref{tab:main_avsd}, M2K-VDG performs significantly better than other methods, particularly on BLEU4, ROUGH-L, and CIDEr metrics. This implies strong fluency and keyword-generation capabilities.
Notably, our method also surpasses MSG-BART, which uses an external scene graph, especially on the CIDEr metric. This shows our methods are capable of generating more significant answers without external input.
Furthermore, the WUPS score in Table~\ref{tab:main_next} surpasses others by a huge margin, which illustrates that our method can improve the capability in semantic representation effectively. 
Also, the results in Table~\ref{tab:main_next} surpass others by a large scale, which illustrates that our method generates more appropriate answers rather than hallucinations.

Furthermore, we compare the detailed capability of different types of questions with the state-of-art model in the MUSIC-AVQA dataset, which is shown in Table~\ref{tab:main_music}. It is important to note that our method outperforms others in almost all question fields, 
which indicates that our method is effective in hallucination reduction overall.
Specifically, the performance in counting questions is particularly outstanding, which may be because the counting question is more related to the video content than others and has more deeper anchor degree in number tokens. Nonetheless, the performance in addressing audio-visual questions, particularly those about location and comparison, falls short of expectations. This underscores the necessity for our method to conscientiously incorporate modality fusion to enhance these specific aspects in the future.
\begin{figure}[t]
\setlength{\belowcaptionskip}{-0.5cm}
    \centering
    \includegraphics[width=0.85\linewidth]{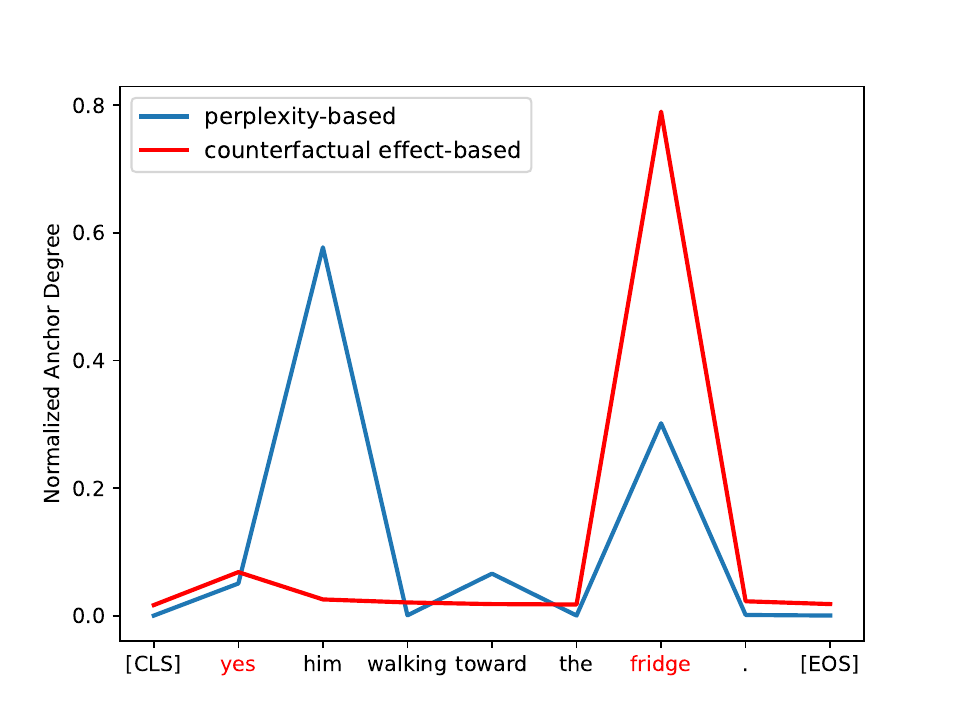}
    \caption{The anchor degree of token derived by two detection methods. The question is `\textbf{is it the last you see of him when he walks out of frame ?}' and the red label denotes the knowledge-related tokens through human detection. 
    We observe that the perplexity-based technique locates `him', `toward', and 'fridge' as the anchor tokens, which should be a set of `yes' and `fridge'. By contrast, the counterfactual effect-based method can detect them accurately,  which demonstrates its effectiveness and robustness.
    }
    \label{fig:ppl-counter}
\end{figure}
\subsection{Ablation Study}
\label{sec:abl}
\subsubsection{Detection Methods}
\label{sec:abl-weight}
To illustrate the effectiveness of the counterfactual effect, we adopt both perplexity- and counterfactual effect-based anchor degree detection methods as described in Section~\ref{sec:weight}, and the results are shown in Table~\ref{tab:abl-formulations} and Figure~\ref{fig:ppl-counter}. As Table~\ref{tab:abl-formulations} shows, both two detection methods can improve the performance and reduce the hallucination across all the metrics, especially on the CIDEr. However, it is noted that the counterfactual effect-based detection strategy enhances the model performance more effectively than the perplexity-based one, which verifies the capability of the counterfactual effect in anchor token detection and hallucination reduction. 
Additionally, Figure~\ref{fig:ppl-counter} also verifies that the counterfactual effect enhances the anchor token detection.
\subsubsection{Baselines Adaptiveness}
To validate the robustness of our method, we extend it to various baseline models and evaluate the performance on the AVSD10 official test set, shown in Table~\ref{tab:abl-methods}. For better validation of the robustness, we conduct experiments on five baseline models that differ in feature extractors, architectures, and modality fusion strategies, which is described in Section~\ref{Sec:baselines}. 
As shown in Table~\ref{tab:abl-methods}, all the baselines show an improvement in all metrics, especially the TSF+M2K can get comparable performance with the ensemble method. Among the metrics, the performance in METEOR, and CIDEr is more effective than others. It is important to note that the improvement in METEOR indicates that our methods can enhance the completeness of information and semantics in responses. Additionally, the improvement in CIDEr further demonstrates that our method can generate more crucial information.


\begin{table}[t]
\setlength{\belowcaptionskip}{-0.5cm}
\centering
\begin{tabular}{lcc}
\hline
\textbf{Anchor}&  \textbf{BLEU4} &  \textbf{CIDEr} \\ \hline
VideoBART     & 0.384  & 0.987   \\
VideoBART+$\operatorname{Q}$  & 0.393   & 1.002   \\
VideoBART+$\operatorname{VA}$    & 0.390  & 1.007   \\
VideoBART+$\operatorname{H}$    & 0.395   & 1.010   \\ 
VideoBART+$\operatorname{K}$    & 0.398  & 1.020   \\ \hline
\end{tabular}
\caption{Evaluation results of different modality ablation experiments on AVSD10 dataset. $\operatorname{VideoBART + X}$ denotes the anchor is established with $\operatorname{X}$.}
\label{tab:abl-modality}
\end{table}

\subsubsection{Modality Anchor}
To further explore the influence of different modalities, we establish a series of experiments on different modality anchor token detection strategies, and the results are shown in Table~\ref{tab:abl-modality}. Specifically, it is important to note that all the different settings lead to performance improvement. Among the settings, the question anchor shows a minimal performance gain which demonstrates that our method needs multimodal knowledge anchor tokens to reduce hallucination. Furthermore, other results also illustrate that the more multimodal knowledge is used in anchor-enhanced generation, the greater performance gains are obtained.

\subsection{Case Study}
\setlength{\belowcaptionskip}{-0.5cm}
\begin{figure}
    \centering
    \includegraphics[width=0.9\linewidth]{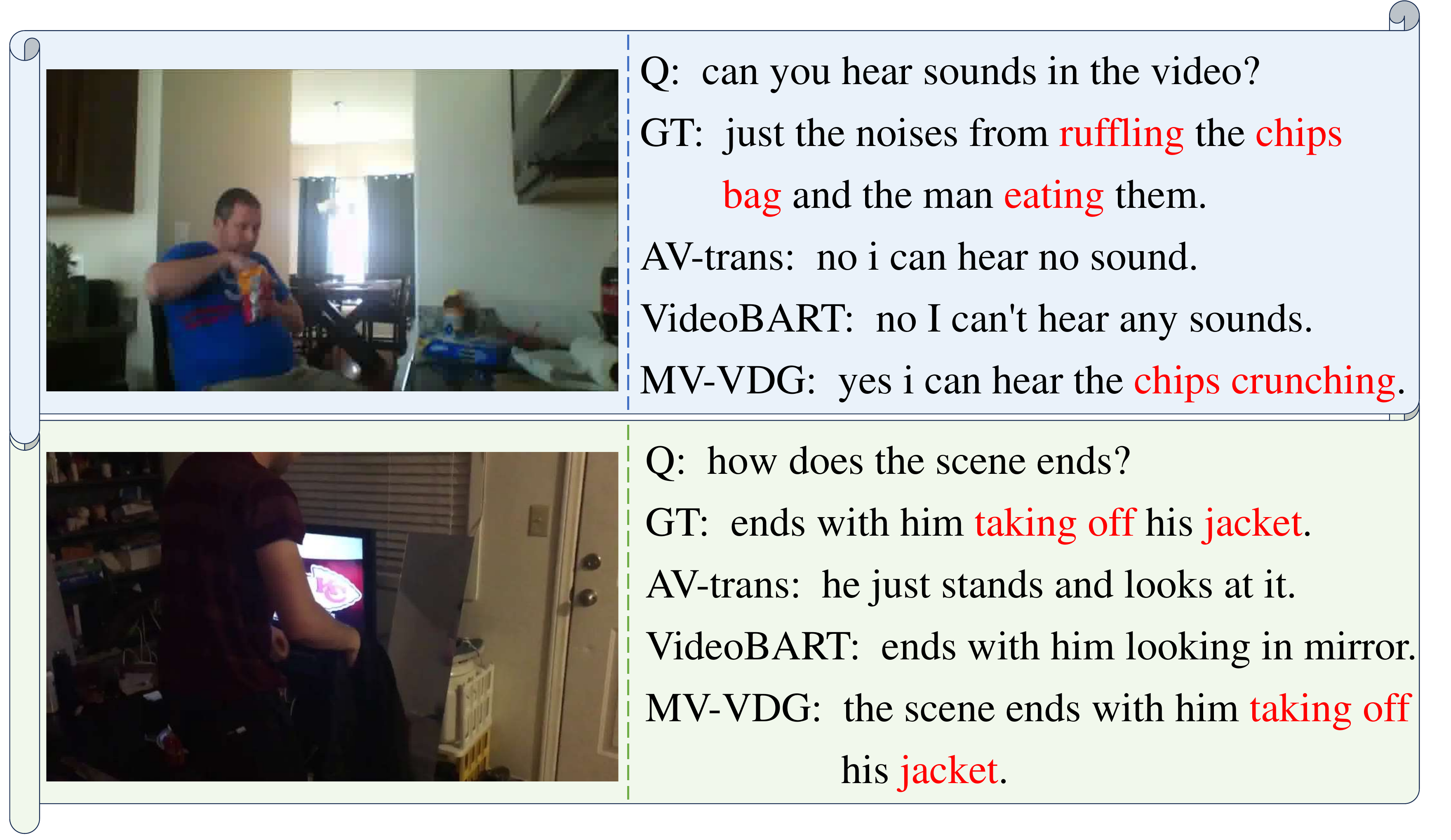}
    \caption{The case study on the AVSD10 dataset, the multimodal knowledge-related token is colored red.}
    \label{fig:case_study1}
\end{figure}

We further examine the long sentence generation ability M2K-VDG on the AVSD dataset, as depicted in Figure~\ref{fig:case_study1}. We consider two examples from the perspectives of auditory and visual hallucination, respectively. The results confirm that our method can efficiently execute the VDG task in both `seeing' and `hearing', particularly in generating multimodal knowledge anchor tokens. This demonstrates its substantial capability in reducing hallucination.

\section{Conclusions}

To reduce the varying hallucinations in the VDG task among different models, we make effective use of the multimodal knowledge anchor tokens in the ground answer, and propose M2K-VDG to achieve model-adaptive hallucination reduction. Specifically, we measure the anchor degree by counterfactual effect and enhance the training by anchor-enhanced loss. Through extensive experiments on three VDG benchmarks, we substantiate the efficacy of our proposed approach in comparison to various state-of-the-art methods.

\section*{Limitations}
Although our model is significantly more effective than the state-of-the-art models, there is much room to improve the performance of anchor token enhancement.
For example, we only consider the anchor tokens for a single specific model but ignore focusing more on the common anchor tokens shared by various models.
Furthermore, both two detection methods are limited to locating the tokens related to specific modalities, but are unable to achieve this goal from other aspects, such as spatial relations and existential questions.
In the future, we will continue to reduce the hallucination by further employing anchor tokens.

\section*{Ethical Consideration}
We aim to propose a system to reduce hallucination by enhancing the multimodal knowledge anchor and provide two methods to detect the anchor tokens precisely. However, the performance of detection can effectively affect the effectiveness of training and there are some risks of potential misuse. For example, the wrong purpose of anchor detection such as violence and pornography scene-anchor tokens detection can lead to an unhealthy multimodal dialogue system. So, we would like to add a protection mechanism to prevent the model from these kinds of words.
\appendix
\label{sec:appendix}
\newpage
\section{Datasets}
\label{sec:appendix1}
\paragraph{AVSD10}AVSD10\footnote{\url{https://github.com/dialogtekgeek/AVSD-DSTC10_Official}} dataset~\citep{avsd} is expanded from the Charades dataset~\cite{charades_dataset} with multi-turns question-answering pairs, which concentrates more on the daily life. There are three relevant datasets: AVSD7, AVSD8, and AVSD10 dataset, but the AVSD10 dataset is the most challenging among these datasets so we only use AVSD10 as the illustration. The questions are proposed to obtain both a comprehensive understanding of the video for overall understanding and a detailed description of objective existence information. As most answers consist of 5-9 words, we use BLEU, METEOR, ROUGE-L, and CIDEr as the validation metrics to evaluate fluency and keyword generation performance.
\paragraph{NExT-OE} NExT-OE dataset\footnote{\url{https://github.com/doc-doc/NExT-OE}}~\citep{xiao2021next} is constructed based on YFCC-100M dataset~\cite{YFCC100M}, which focus more on outdoor activities. The questions focus on visual reasoning ability, which consists of casualty, temporary reasoning, and descriptive ability, and the answers are shorter than 4 words with certain semantics. Considering this, we use the WUPS score to evaluate the semantic similarity between the answer and the ground truth.
\paragraph{MUSIC-AVQA} MUSIC-AVQA dataset\footnote{\url{https://gewu-lab.github.io/MUSIC-AVQA/}}~\citep{Li_2022_CVPR} is expanded from musical performance both real videos and synthesis videos. The questions are more attentive to the audio reasoning, and the visual reasoning questions are also contained.  As each question has a certain answer with 1 word, we use accuracy as the evaluation metric.
\newpage
\bibliography{anthology,custom}
\bibliographystyle{acl_natbib}

\end{document}